\documentclass{article}

\usepackage[numbers]{natbib} 
\usepackage{authblk}         
\usepackage{graphicx}        
\usepackage{hyperref}        
\usepackage{arxiv}
\usepackage{fancyhdr}
\usepackage{acronym}
\usepackage{amsmath}
\usepackage{algorithm}
\usepackage{algorithmic}
\usepackage{subcaption}
\usepackage[inline]{enumitem}
\usepackage[utf8]{inputenc} 
\usepackage[T1]{fontenc}    
\usepackage{url}     
\usepackage{amsfonts}       
\usepackage{nicefrac}      
\usepackage{microtype}      
\usepackage{lipsum}		
\usepackage{graphicx}
\usepackage{doi}
\usepackage{graphicx} 
\usepackage{adjustbox}
\usepackage{adjustbox} 
\usepackage{float} 
\usepackage[hypcap=true]{caption}
\usepackage{adjustbox}
\usepackage{booktabs}
\usepackage{placeins} 
\usepackage{colortbl}   
\usepackage[table,xcdraw]{xcolor}

\title{IBO: Inpainting-Based Occlusion to Enhance Explainable Artificial Intelligence Evaluation in Histopathology}

\newcommand*{\affaddr}[1]{#1} 
\newcommand*{\affmark}[1][*]{\textsuperscript{#1}}
\newcommand*{\email}[1]{\texttt{#1}}

\author{%
Pardis Afshar\affmark[1]\href{https://orcid.org/0009-0001-1825-8603}{\includegraphics[scale=0.06]{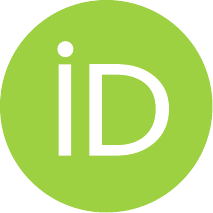}},
Sajjad Hashembeiki\affmark[2]\href{https://orcid.org/0009-0004-3163-4302}{\includegraphics[scale=0.06]{orcid.pdf}},
Pouya Khani\affmark[3]\href{https://orcid.org/0000-0002-4786-0572}{\includegraphics[scale=0.06]{orcid.pdf}},\\
Emad Fatemizadeh\affmark[2]\href{https://orcid.org/0000-0001-5017-7136}{\includegraphics[scale=0.06]{orcid.pdf}},
and Mohammad Hossein Rohban\affmark[1]\href{https://orcid.org/0000-0001-6589-850X}{\includegraphics[scale=0.06]{orcid.pdf}}\\
\affaddr{\affmark[1]Department of Computer Engineering, Sharif University of Technology, Tehran, Iran}\\
\affaddr{\affmark[2]Department of Electrical Engineering, Sharif University of Technology, Tehran, Iran}\\
\affaddr{\affmark[3]Department of Computer Science, Aarhus University, Aarhus, Denmark}\\
\email{\{pafshar, s.hashem, fatemizadeh, rohban\}@sharif.edu}\\
\email{pouya.khani@cs.au.dk}%
}

\renewcommand{\shorttitle}{IBO: Inpainting-Based Occlusion}

\hypersetup{
pdftitle={IBO: Inpainting-Based Occlusion to Enhance Explainable Artificial Intelligence Evaluation in Histopathology},
pdfsubject={cs.cs},
pdfauthor={Pardis Afshar, Sajjad Hashembeiki, Pouya Khani, Emad Fatemizadeh, Mohammad Hossein Rohban},
pdfkeywords={Histopathology, Histopathological Image Analysis, Cancer Diagnosis, Deep Diffusion Models, Explainable Artificial Intelligence},
}

\begin{document}
\maketitle

\begin{abstract}
Histopathological image analysis is crucial for accurate cancer diagnosis and treatment planning. While deep learning models, especially convolutional neural networks, have advanced this field, their "black-box" nature raises concerns about interpretability and trustworthiness. Explainable Artificial Intelligence (XAI) techniques aim to address these concerns, but evaluating their effectiveness remains challenging. A significant issue with current occlusion-based XAI methods is that they often generate Out-of-Distribution (OoD) samples, leading to inaccurate evaluations. In this paper, we introduce Inpainting-Based Occlusion (IBO), a novel occlusion strategy that utilizes a Denoising Diffusion Probabilistic Model to inpaint occluded regions in histopathological images. By replacing cancerous areas with realistic, non-cancerous tissue, IBO minimizes OoD artifacts and preserves data integrity. We evaluate our method on the CAMELYON16 dataset through two phases: first, by assessing perceptual similarity using the Learned Perceptual Image Patch Similarity (LPIPS) metric, and second, by quantifying the impact on model predictions through Area Under the Curve (AUC) analysis. Our results demonstrate that IBO significantly improves perceptual fidelity, achieving nearly twice the improvement in LPIPS scores compared to the best existing occlusion strategy. Additionally, IBO increased the precision of XAI performance prediction from 42\% to 71\% compared to traditional methods. These results demonstrate IBO's potential to provide more reliable evaluations of XAI techniques, benefiting histopathology and other applications. The source code for this study is available on GitHub\footnote{\url{https://github.com/a-fsh-r/IBO}}.
\end{abstract}


\section{Introduction}
Histopathological image analysis plays a critical role in diagnosing and planning the treatment of cancers, such as breast cancer \cite{7900002}. Deep learning has significantly advanced the field of breast cancer histopathological image classification \cite{dimitriou2019deep}. Traditional methods, which relied on extensive manual feature engineering and domain expertise, have been outperformed by modern techniques \cite{beyer2023human}. \ac{CNN} architectures, in particular, automate the process of feature extraction and classification directly from raw image data. This automation simplifies the workflow and enhances accuracy, providing a powerful tool in the diagnosis of breast cancer \cite{9269335}.

Despite deep models' high performance, the \textit{black-box} nature of these models raises concerns about their interpretability and transparency \cite{lucieri2020achievements}. Pathologists and experts need models that not only perform exceptionally well but also clearly explain their decision-making processes \cite{https://doi.org/10.1002/widm.1474}. To address this, \ac{XAI} techniques are being increasingly integrated for explaining machine learning models in various fields, such as computer networking \cite{khani2024explainable}, indoor image classification \cite{aminimehr2024tbexplain}, brain-computer interfaces \cite{rabiee2024wavelet}, and especially for our case, into histopathological cancer diagnosis \cite{Tosun2020}. \ac{XAI} makes the decision-making processes of complex models, such as \ac{CNN}s, understandable to humans. This transparency is crucial in the medical field, where understanding how \ac{AI} reaches its conclusions can help in clinical validation and build trust among healthcare professionals \cite{VANDERVELDEN2022102470}.

However, evaluating the reliability of different \ac{XAI} methods remains challenging \cite{lin2021you}. This challenge highlights the urgent need for effective evaluation frameworks. Given that experts have limited time for manual assessments, automating the evaluation process is essential for ensuring efficient and timely evaluations \cite{Bu_inca_2020}. Evaluation metrics for \ac{XAI} methods are typically divided into two categories: qualitative and quantitative. Qualitative metrics, which rely on subjective assessments, contrast sharply with quantitative metrics that offer objective, numerical evaluations. The latter are particularly valuable for their reproducibility, scalability, and their ability to facilitate systematic tuning of models. One major challenge in \ac{XAI} evaluation is the absence of ground-truth explanations, making it difficult to objectively assess the accuracy and reliability of the explanations generated by different methods. Without a definitive standard, comparing and validating the effectiveness of \ac{XAI} techniques becomes inherently subjective and complex. In cases where ground-truth is not available, alternative quantitative evaluation methods can be utilized. 

Quantitative metrics assess various critical aspects of \ac{XAI}, such as trustworthiness (often measured as fidelity to the original model's decisions) \cite{tomsett2020sanity}, sensitivity \cite{yeh2019infidelitysensitivityexplanations} to changes in input, and robustness against variations in data, etc. Among these, fidelity is considered one of the most crucial, as it directly measures how faithfully an explanation represents the decision-making process of the underlying model. To evaluate fidelity, researchers have developed several classes of quantitative metrics, including perturbation-based and occlusion-based metrics \cite{zeiler2014visualizing}. Occlusion-based metrics are among the most well-known ones due to their straightforward implementation and insightful evaluations. These metrics systematically mask input data to observe how changes affect model outputs, thereby providing a clear window into the model’s operational dependencies. Using these metrics, the fidelity of \ac{XAI} methods can be evaluated by assessing how rapidly the model's performance degrades when portions of the input are masked \cite{IVANOVS2021228}.

Existing occlusion strategies in \ac{XAI} evaluation metrics, such as the Deletion method \cite{samek2016evaluating}, face significant challenges that undermine their effectiveness. A primary issue arises when these methods attempt to mask or occlude input features, resulting in the transformation of the input sample into an \ac{OoD} or anomalous sample \cite{mirónicolau2024comprehensivestudyfidelitymetrics}. This transformation introduces ambiguity in the model's behavior change, as it becomes unclear whether the observed change is due to the occlusion itself, the \ac{OoD} nature, or the anomaly of the modified sample \cite{gomez2022metricssaliencymapevaluation}. Consequently, the accuracy and reliability of these occlusion-based metrics are compromised. To address this, there is a critical need for more sophisticated occlusion strategies that minimize the \ac{OoD} ratio and the introduction of anomalies in occluded samples. Such strategies should ensure that the modified inputs remain within the distribution of the training data, thereby providing a more precise and faithful assessment of the model's reliance on specific input features. By reducing the incidence of \ac{OoD} and anomalous samples, we can achieve a more accurate understanding of the model's decision-making process and improve the overall robustness of \ac{XAI} evaluation metrics.

In this paper, we propose a novel occlusion strategy that leverages a diffusion model trained on non-cancerous samples of histopathological images. Our approach involves inpainting cancerous regions with non-cancerous regions as part of a region occlusion process to remove information during the evaluation of \ac{XAI} methods. By providing more realistic inpainting of occluded regions, our method aims to overcome the limitations of existing techniques, reducing \ac{OoD} artifacts and anomalies while preserving the integrity of the original data distribution. This framework enhances the reliability of the explanation evaluation process, offering a more trustworthy and efficient explanation method for histopathological image diagnosis.

The reminder of this paper is structured as follows: Section \ref{related} reviews related work, highlighting previous approaches and their limitations. In Section \ref{proposed}, we detail our proposed methodology, including the occlusion strategy using inpainting approach. Section \ref{experiments} presents our experiments and results, demonstrating the effectiveness of our proposed method. Section \ref{future} discusses the limitations of our study and suggests directions for future work. Finally, Section \ref{Conclusion} concludes the paper, summarizing our key findings and contributions to the field of Explainable histopathology images diagnosis and also \ac{XAI} methods evaluation.

\section{Related works}
\label{related}

In this section, we highlight that our study closely aligns with two main areas of existing research. Firstly, it pertains to the evaluation of explanation methods, particularly within the realm of medical imaging and specifically histopathology. Our work contributes to this field by providing a more precise evaluation framework, enabling more accurate and reliable comparisons of existing \ac{XAI} methods. This, in turn, allows us to select the most effective methods for our specific case within the domain of medical imaging. Secondly, our study is related to occlusion strategies within quantitative occlusion-based evaluation metrics. In the following subsections, we will review each of these areas in detail, discussing the most relevant and closely related previous works.

\subsection{Explainable AI evaluation}

Several studies have compared \ac{XAI} evaluation metrics to assess their effectiveness and reliability. 
Kindermans et al. \cite{kindermans2019unreliability} highlighted the (un)reliability of saliency methods, emphasizing the need for robust evaluation frameworks across domains including medical images. 
Dasgupta et al. \cite{dasgupta2022framework} proposed a framework for evaluating the faithfulness of local explanations, further contributing to the comparative assessment of \ac{XAI} methods. 
Nauta et al. \cite{nauta2023systematic} conducted a systematic review on quantitative evaluation methods, providing a detailed comparison across multiple \ac{XAI} techniques. 
These studies collectively underscore the importance of systematic and rigorous evaluation metrics in the broader \ac{XAI} landscape.

In the context of medical imaging, several studies have specifically addressed the evaluation of \ac{XAI} methods. 
Lamprou et al. \cite{lamprou2024evaluation} focused on evaluating deep learning interpretability methods for medical images under the scope of faithfulness, providing insights into their applicability and reliability in clinical settings. Furthermore, studies by Rajaraman et al. \cite{rajaraman2019visualizing} and Sayres et al. \cite{sayres2019using} explored the use of deep learning models and integrated gradients \cite{sundararajan2017axiomaticattributiondeepnetworks} explanations in detecting diseases from medical images, demonstrating practical applications and evaluation of \ac{XAI} methods.

Specifically, for histopathological images, evaluating the effectiveness of \ac{XAI} methods remains a critical challenge. 
Graziani et al. \cite{graziani2021evaluation} provided a comprehensive evaluation of \ac{CNN} visual explanations in histopathology, offering a comparative analysis of different \ac{XAI} techniques. Their findings underscore the necessity for domain-specific evaluation metrics to accurately assess and compare interpretability methods in histopathological image analysis.

By integrating these studies, our work contributes to the field by proposing a more precise evaluation framework tailored to medical imaging, particularly histopathology. This enables more accurate comparisons of existing \ac{XAI} methods, facilitating the selection of the most effective techniques for specific clinical applications.

\subsection{Occlusion strategy}

In the evaluation of \ac{XAI} methods, various strategies have been proposed to occlude features and assess their importance. Blackening \cite{tomsett2019sanitycheckssaliencymetrics} involves occluding features by replacing them with black pixels (zero intensity). This straightforward method helps in identifying the importance of specific regions in an image. Given an image \( I \) and a binary mask \( M \) indicating the regions to be occluded, the blackened image \( I' \) is defined as:
\begin{equation}
I' = I \odot (1 - M)
\end{equation}
where \( \odot \) denotes element-wise multiplication. While blackening is easy to implement, it often introduces \ac{OoD} artifacts, leading to unrealistic samples that do not accurately represent the original data distribution. Another strategy is Blurring \cite{Fong_2017} that replaces the occluded regions with a Gaussian blur, which maintains the overall structure of the image while removing fine details. Let \( G \) be a Gaussian filter, the blurred image \( I' \) is given by:
\begin{equation}
I' = I \odot (1 - M) + (I * G) \odot M
\end{equation}
where \( \odot \) denotes element-wise multiplication and \( * \) denotes convolution. Blurring reduces \ac{OoD} artifacts compared to blackening by keeping occluded regions contextually similar to their surroundings. However, when this strategy is applied, the features are not completely deleted or removed due to the blurring effect, which only obscures but does not eliminate the information. In this study \cite{tomsett2019sanitycheckssaliencymetrics}, they replaces occluded regions with the mean pixel value of the dataset, offering a neutral replacement. Let \( \mu \) be the mean pixel value, then the occluded image \( I' \) is:
\begin{equation}
I' = I \odot (1 - M) + \mu \odot M
\end{equation}

where \( \odot \) denotes element-wise multiplication. This method balances maintaining data integrity and avoiding extreme \ac{OoD} artifacts, providing a middle ground between blackening and blurring. In the Histogram strategy \cite{8546302}, they replaced occluded regions with values sampled from the histogram of pixel intensities of the image, maintaining the statistical properties of the original image. For each occluded pixel in \( P \), sample a value from the histogram \( H \) of \( I \):
\begin{equation}
I' = I \odot (1 - M) + H \odot M
\end{equation}
where \( \odot \) denotes element-wise multiplication. Histogram-based occlusion preserves local image statistics better than mean replacement and provides a more realistic replacement than blackening or blurring, though it can still introduce some artifacts. Finally, \ac{NLI} \cite{rong2022consistent} occludes features by replacing them with values derived from a linear combination of the surrounding observed features, introducing noise to simulate realistic scenarios. This method preserves the local dependencies and adds variability to prevent the model from relying on occlusion artifacts. Given a feature vector \( \mathbf{x} \) with a mask \( \mathbf{m} \) indicating the occluded regions, the imputed feature vector \( \mathbf{x}' \) is computed as:

\begin{equation}
\mathbf{x}' = (1 - \mathbf{m}) \odot \mathbf{x} + \mathbf{m} \odot (\mathbf{X} \hat{\beta} + \epsilon)
\end{equation}

where \( \odot \) denotes element-wise multiplication. Here, \( \mathbf{X} \) is the matrix of observed features, \( \hat{\beta} \) is the vector of coefficients estimated from a linear model fit to the observed data, and \( \epsilon \) is a noise term sampled from a normal distribution \( \mathcal{N}(0, \sigma^2) \). This approach helps in maintaining the integrity of the model evaluation by ensuring that the imputed features are both realistic and varied, thus providing a more robust assessment of feature importance in \ac{XAI} tasks. However, the occluded regions remain noticeably different from the rest of the image due to the blurriness.

\section{Methodology}
\label{proposed}

\begin{figure}[t]
    \centering
    \includegraphics[width=1\textwidth]{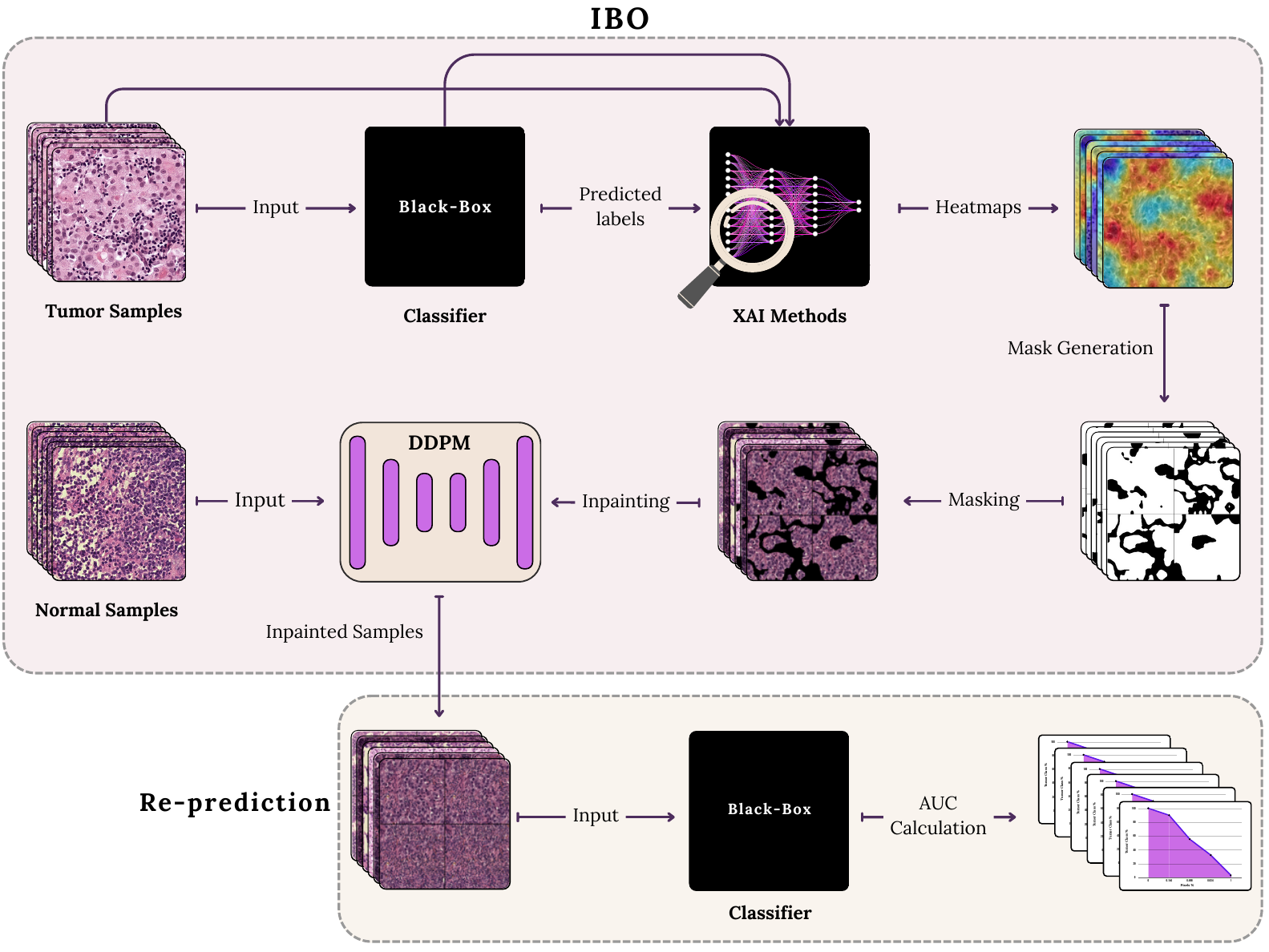}
    \caption{Overview of the methodology used for evaluating \ac{XAI} methods for breast cancer diagnosis from histopathological images. In Masks, black regions indicate the areas that should be inpainted. These black regions correspond to the tumoric areas identified by the \ac{XAI} approaches, which are crucial for evaluating the effectiveness of our methodology.}
    \label{fig:methodology}
\end{figure}

The methodology is outlined in the following steps, as illustrated in Figure \ref{fig:methodology}. Our \ac{IBO} proposed framework involves generating heatmaps to identify tumoric regions, creating corresponding masks, and then masking the patches based on these masks. We also trained a \ac{DDPM} \cite{ho2020denoisingdiffusionprobabilisticmodels} on normal samples to learn the distribution and features of normal tissues. This trained \ac{DDPM} is then used to inpaint the masked regions, effectively replacing tumoric areas with features typical of normal tissues. Subsequently, we re-predict the inpainted samples using our classifier to predict tumor probabilities. The final step is to calculate the \ac{AUC} to assess the impact of the inpainted regions.

\subsection{Inpainting}

In our study, we utilized RePaint \cite{lugmayr2022repaintinpaintingusingdenoising}, an approach that leverages an unconditional \ac{DDPM} trained exclusively on normal tissue patches to learn the distribution of normal features. This technique allows us to inpaint regions identified as tumorous in histopathological images, effectively replacing these areas with realistic, normal tissue content. RePaint capitalizes on the \ac{DDPM}’s capability to generate high-resolution images while maintaining semantic consistency and processing efficiency.

A significant advantage of RePaint is its versatility; it can handle various mask distributions without requiring additional training on specific masks. This adaptability is particularly effective for evaluating and interpreting the significance of features highlighted by different \ac{CAM} \cite{Oquab_2015_CVPR} techniques, as each \ac{XAI} approach may produce heatmaps and masks with differing distributions. This variability in mask distributions means the method remains robust across different \ac{XAI} approaches, ensuring a comprehensive evaluation of feature importance. By inpainting tumorous areas, we can assess the impact of removing these features on the model’s predictions, ensuring the inpainted regions blend seamlessly with their surroundings. However, while RePaint can generate textures that match neighboring areas, the inpainted regions may sometimes lack semantic harmony with the rest of the image. To address this, a novel resampling approach was implemented to improve the semantic consistency of these regions \cite{lugmayr2022repaintinpaintingusingdenoising}. This approach involves applying forward transitions to diffuse the output from $x_{t-1}$ back to $x_t$, scaling back the image and adding noise to preserve information from the generated regions. The generated regions $x^{\text{unknown}}_{t}$ are then updated to be more consistent with the known regions $x^{\text{known}}_{t}$.

The resampling operation works incrementally, harmonizing one step at a time, which might not fully integrate semantic information across the entire denoising process. To manage this, a jump length $j$ is defined \cite{lugmayr2022repaintinpaintingusingdenoising}, determining the number of forward transitions before applying reverse transitions. In this study, both the jump length and the number of resamplings are set to 10, striking a balance between image quality and computational efficiency, ensuring enhanced semantic consistency and manageable processing time. In Algorithm \ref{algo1}, the procedure of RePaint method is shown.

\begin{algorithm}
\caption{RePaint Algorithm}
\begin{algorithmic}[1]
\REQUIRE Image $x$, Binary mask $m$ where 0
indicates region to be inpainted, Total diffusion steps $T$, Number of resampling steps $U$
\ENSURE Inpainted image $x_0$
\STATE $x_T \sim \mathcal{N}(0, I)$
\FOR{$t = T$ \textbf{to} $1$}
    \FOR{$u = 1$ \textbf{to} $U$}
        \STATE $\epsilon \sim \mathcal{N}(0, I)$ \textbf{if} $t > 1$, \textbf{else} $\epsilon = 0$
        \STATE $x_{t-1}^{\text{known}} = \sqrt{\bar{\alpha}_t} x_0 + (1 - \bar{\alpha}_t) \epsilon$
        \STATE $z \sim \mathcal{N}(0, I)$ \textbf{if} $t > 1$, \textbf{else} $z = 0$
        \STATE $x_{t-1}^{\text{unknown}} = \frac{1}{\sqrt{\alpha_t}} \left( x_t - \frac{\beta_t}{\sqrt{1 - \bar{\alpha}_t}} \epsilon_\theta(x_t, t) \right) + \sigma_t z$
        \STATE $x_{t-1} = m \odot x_{t-1}^{\text{known}} + (1 - m) \odot x_{t-1}^{\text{unknown}}$
        \IF{$u < U$ \textbf{and} $t > 1$}
            \STATE $x_t \sim \mathcal{N}\left( \sqrt{1 - \beta_{t-1}} x_{t-1}, \beta_{t-1} I \right)$
        \ENDIF
    \ENDFOR
\ENDFOR
\RETURN $x_0$
\end{algorithmic}
\label{algo1}
\end{algorithm}

Where $T$ denotes the total number of diffusion steps, $U$ is the number of resampling steps at each time step, $m$ represents the mask matrix, and $\odot$ signifies element-wise multiplication. Note that a jump length of $j = 1$ is used in the algorithm above.

\subsection{Masking}

To support the inpainting process, a clustering approach is employed to segment heatmaps into regions of interest. This segmentation aids in guiding the inpainting procedure by highlighting areas identified as significant by \ac{CAM} methods. K-means clustering \cite{Hartigan1979} is applied to the grayscale pixel values of each heatmap, which are based on pixel intensity, to partition these pixels into \( k \) clusters, with \( k = 5 \) in this study. The objective of clustering is to minimize the \ac{WCSS}, which is defined as:

\begin{equation}
\text{\ac{WCSS}} = \sum_{i=1}^k \sum_{x_j \in C_i} \| x_j - \mu_i \|^2
\end{equation}

Here, \( \mu_i \) represents the centroid of cluster \( C_i \), and \( x_j \) is a pixel in that cluster. The algorithm iterates to optimize the centroids and pixel assignments, effectively clustering the grayscale values based on intensity into distinct groups.

In this study, the inpainting process is conducted iteratively, focusing on different regions of the image based on their importance, as indicated by the color intensity of the heatmap. With \(k = 5\), the heatmap is divided into five distinct masks, each corresponding to regions of varying significance:
    \begin{enumerate*}[label=\Roman*.]
        \item Red Areas: Most critical regions for model decision-making,
        \item Red to Yellow Areas: High significance regions,
        \item Yellow to Green Areas: Moderately important regions,
        \item Green to Light Blue Areas: Regions with lower importance,
        \item Light Blue to Blue Areas: Least significant regions, often considered non-essential for decision-making and excluded from the inpainting process. 
    \end{enumerate*}
    
    \begin{figure*}
    \centering
    \begin{subfigure}[t]{0.25\textwidth}
        \centering
        \includegraphics[width=\linewidth]{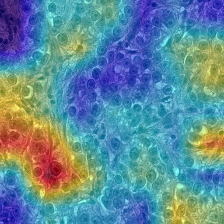} 
        \caption{Generated heatmap}
        \label{fig:heatmap}
    \end{subfigure}
    \hspace{0.05\textwidth}
    \begin{subfigure}[t]{0.25\textwidth}
        \centering
        \includegraphics[width=\linewidth]{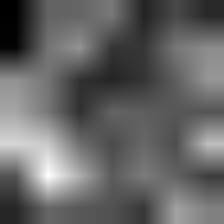} 
        \caption{Grayscale version of the heatmap}
        \label{fig:grayscale}
    \end{subfigure}
    \hspace{0.05\textwidth}
    \begin{subfigure}[t]{0.25\textwidth}
        \centering
        \includegraphics[width=\linewidth]{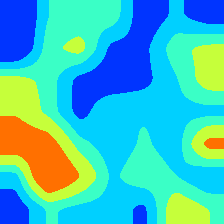} 
        \caption{Levels of importance in the heatmap}
        \label{fig:level}
    \end{subfigure}
    \caption{Comparison of generated heatmap, grayscale version of the heatmap, and levels of importance.}
    \label{fig:combined}
\end{figure*}

Figures \ref{fig:heatmap}, \ref{fig:grayscale}, and \ref{fig:level} illustrate the generated heatmap, its grayscale version, and the corresponding significance levels. During each inpainting step, only one specific mask (for example, red to yellow areas) is targeted for inpainting. This means that regions corresponding to previously inpainted masks remain unchanged in subsequent steps. The process is cumulative, where the output image from each step is carried forward and further inpainted based on the next mask. As a result, the final inpainted image reflects progressive modifications, with the more critical areas being addressed first, while less significant regions remain unmodified.

\subsection{Re-prediction}

After the inpainting process, the modified images are reintroduced into the trained classifier for re-prediction, focusing on the classification of each patch. This re-prediction is crucial in determining whether the probability of tumor classification shifts following the removal of the highlighted features. By comparing the classifier’s predictions before and after inpainting, we can critically assess the impact of the identified features on the model's decision-making process. Furthermore, this approach enables us to evaluate the effectiveness of different \ac{XAI} methods. The accuracy of each \ac{XAI} method in pinpointing tumoric regions is measured by observing changes in the model's predictions, thereby identifying the most reliable and informative heatmaps for feature interpretation.

\section{Experiments}
\label{experiments}
In this section, we provide a comprehensive evaluation of our proposed methodology.
\subsection{Configuration}

\begin{figure}[h]
    \centering
    \includegraphics[width=.9\textwidth]{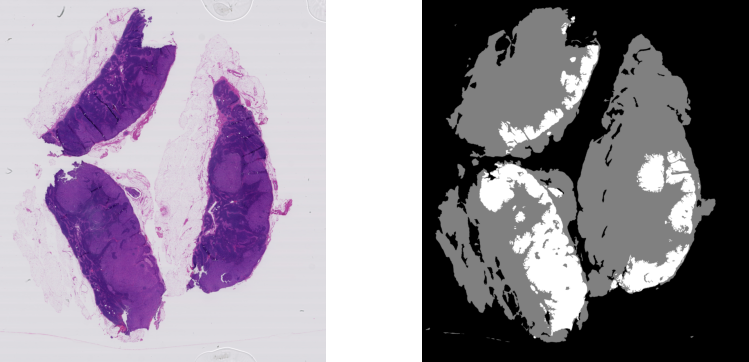}
    \caption{Side-by-side comparison of the \ac{WSI} (left) and the mask image (right). In the mask image, the white regions indicate tumoric areas.}
    \label{fig:camelyon}
\end{figure}

CAMELYON16 dataset \cite{Ehteshami2017Bejnordi} is a widely recognized benchmark in the field of cancer classification. In our study, we leverage this dataset, which comprises 400 \ac{WSI} of sentinel lymph nodes, including 270 slides with precise pixel-level annotations. We further divide these annotated slides into training, validation, and test sets to facilitate our analysis. The dataset provides ground truth data in two formats: \ac{XML} files with contour annotations and \ac{WSI} binary masks. To evaluate \ac{XAI} methods, we use the \ac{WSI} binary masks to calculate the \ac{IoU} by comparing the masks with the heatmaps generated by these methods. The visual representation of one \ac{WSI} from the CAMELYON16 dataset is displayed in Figure \ref{fig:camelyon}.

As part of the data pre-processing phase, we extract \(512 \times 512\) pixel patches from each \ac{WSI} and its corresponding binary mask. This approach ensures that each patch is manageable in size while retaining sufficient detail for analysis. To maintain consistency across histopathological images, we employ stain normalization using the Modified Reinhard method \cite{9616117}. This technique corrects for variations in staining between different tissue samples and slide preparations by aligning the color distribution of the input images to a reference stain color distribution. By applying stain normalization, we reduce the impact of staining variability on model performance, leading to more accurate and reliable analysis of histopathological images. This step is crucial for improving the consistency and effectiveness of classification and \ac{XAI} methods.

We trained a ResNet50 \cite{he2015deepresiduallearningimage} model on a dataset of tumor and normal samples extracted from the CAMELYON16 dataset, achieving an impressive overall accuracy of 98.62\%. The model demonstrated strong performance with an F1-score of 98.6\% in identifying tumor samples, showcasing its high efficacy in distinguishing between tumor and normal tissue. The classifier was trained for 20 epochs using 2 × T4 GPUs on the Kaggle platform.

In addition to training the ResNet50 model, we also developed an unconditional \ac{DDPM} using normal tissue samples from the CAMELYON16 dataset. This training aimed to generate realistic normal tissue samples for data inpainting. The \ac{DDPM} was trained for 140 epochs on an NVIDIA A100 GPU, ensuring the computational efficiency needed for large-scale histopathological data.

We chose the \ac{DDPM} for its ability to model complex data distributions and produce high-fidelity samples. By training exclusively on normal samples, the \ac{DDPM} learned the underlying distribution of non-tumorous tissues, which is useful for inpainting tasks. This approach is especially valuable when dealing with missing or corrupted image regions, as the \ac{DDPM} can generate plausible tissue structures to fill in the masked areas.

\begin{figure*}[h]
    \centering
    \begin{subfigure}[b]{0.18\textwidth}
        \centering
        \includegraphics[width=\textwidth]{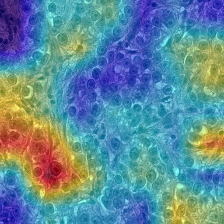}
        \caption{\scriptsize Grad-CAM}
        \label{fig:image_1}
    \end{subfigure}\hspace{0.5em} 
    \begin{subfigure}[b]{0.18\textwidth} 
        \centering
        \includegraphics[width=\textwidth]{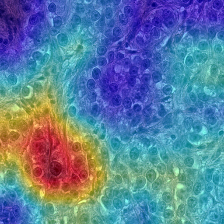}
        \caption{\scriptsize Grad-CAM++}
        \label{fig:image_2}
    \end{subfigure}\hspace{0.5em} 
    \begin{subfigure}[b]{0.18\textwidth} 
        \centering
        \includegraphics[width=\textwidth]{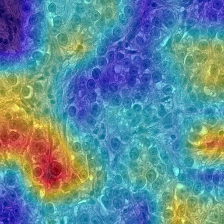}
        \caption{\scriptsize XGrad-CAM}
        \label{fig:image_3}
    \end{subfigure}\hspace{0.5em} 
    \begin{subfigure}[b]{0.18\textwidth} 
        \centering
        \includegraphics[width=\textwidth]{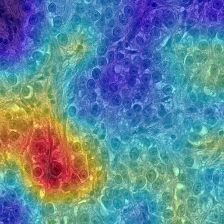}
        \caption{\scriptsize Ablation-CAM}
        \label{fig:image_4}
    \end{subfigure}
    \\ 
    \begin{subfigure}[b]{0.18\textwidth} 
        \centering
        \includegraphics[width=\textwidth]{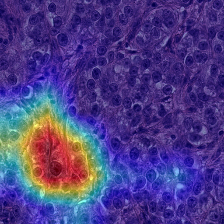}
        \caption{\scriptsize Eigen-CAM}
        \label{fig:image_5}
    \end{subfigure}\hspace{0.5em} 
    \begin{subfigure}[b]{0.18\textwidth}
        \centering
        \includegraphics[width=\textwidth]{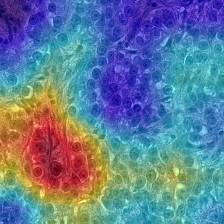}
        \caption{\scriptsize Score-CAM}
        \label{fig:image_6}
    \end{subfigure}\hspace{0.5em} 
    \begin{subfigure}[b]{0.18\textwidth}
        \centering
        \includegraphics[width=\textwidth]{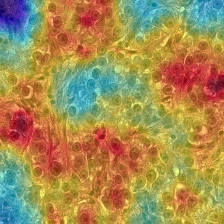}
        \caption{\scriptsize Full-Grad}
        \label{fig:image_7}
    \end{subfigure}
    \caption{A sequence of images showcasing different \ac{CAM}-based methods applied to the tumor patch for decision making. Each method provides a unique perspective on the activation areas within the image.}
    \label{fig:smaller_images}
\end{figure*}

To evaluate the performance of various explanation methods, we applied several \ac{CAM}-based \ac{XAI} techniques, including Grad-CAM \cite{Selvaraju_2019}, Grad-CAM++ \cite{8354201}, XGrad-CAM \cite{fu2020axiombasedgradcamaccuratevisualization}, Ablation-CAM \cite{9093360}, Eigen-CAM \cite{Muhammad_2020}, Score-CAM \cite{wang2020scorecamscoreweightedvisualexplanations}, and Full-Grad \cite{srinivas2019fullgradientrepresentationneuralnetwork}. For each tumoric sample, these techniques generated seven different heatmaps. Figure \ref{fig:smaller_images} displays the heatmaps generated for a tumoric breast histopathological sample. These heatmaps were then used to create masks, which were applied to various occlusion methods, including our \ac{IBO} approach.

\begin{figure}[h]
    \centering
    \includegraphics[width=\textwidth]{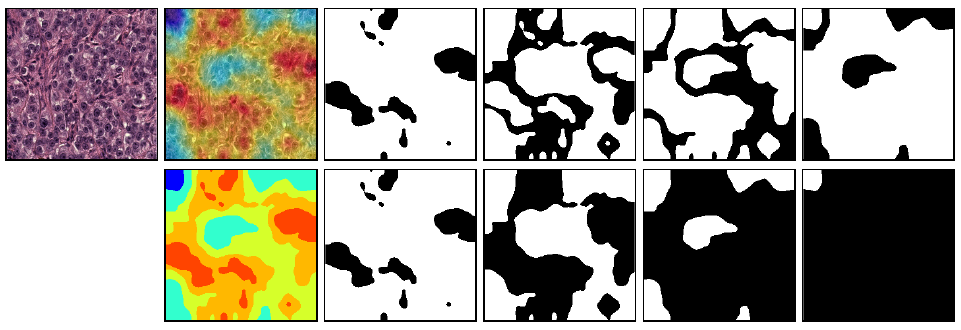}
    \caption{A sequence showing a tumor patch, its Full-Grad heatmap, and various masks. Line 1 displays the original tumor patch and its corresponding heatmap, with masks highlighting regions of varying importance from high to low. Line 2 illustrates that the black regions in each mask represent areas that have been occluded after each step. This sequence demonstrates the progressive occlusion of tumor regions based on their significance as indicated by the heatmap.}

    \label{sample}
\end{figure}

\begin{figure}[t] 
    \centering
    \includegraphics[width=\textwidth]{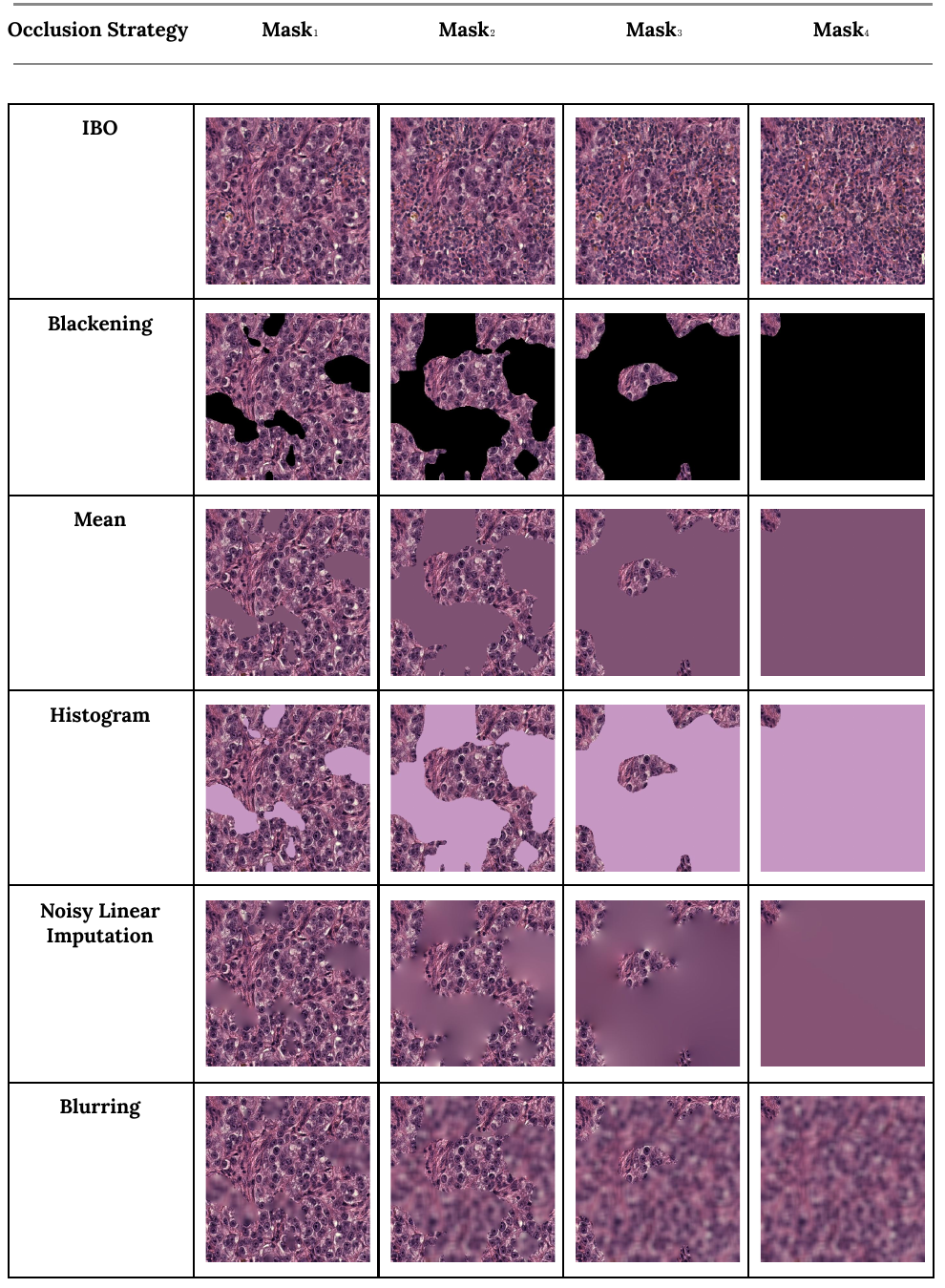}
    \caption{Illustration of various occlusion strategies applied to masked patches.}
    \label{Occlusion_strategy}
\end{figure}

For performance evaluation, we randomly selected 100 samples from the Tumoric test set. Heatmaps and masks were generated for each sample, and various occlusion strategies, including our proposed method, were applied to evaluate their effectiveness. To handle the computational demands of the \ac{IBO} approach, we utilized NVIDIA A100 GPUs. Figure \ref{sample} displays the original patch, the generated heatmap, and the corresponding masks. Figure \ref{Occlusion_strategy} illustrates the different occlusion strategies applied to the patch.

\subsection{Evaluation of Inpainted Samples}

The \ac{LPIPS} metric \cite{zhang2018unreasonableeffectivenessdeepfeatures}, is a widely recognized and effective measure for capturing perceptual differences between images. \ac{LPIPS} is particularly well-suited for assessing image quality and dissimilarities, as it assigns higher values to pairs of images that are perceived as more dissimilar, thereby providing a robust evaluation of perceptual differences.

The \ac{LPIPS} metric is defined as:
\begin{equation}
\text{\ac{LPIPS}}(x, \hat{x}) = \sum_{l} \frac{1}{H_l W_l} \sum_{h=1}^{H_l} \sum_{w=1}^{W_l} \|\mathbf{w}_l \odot (\phi_l(x)_{hw} - \phi_l(\hat{x})_{hw})\|_2^2,
\end{equation}
where \(x\) and \(\hat{x}\) are the original and reconstructed images, respectively, \(\phi_l(\cdot)\) represents the deep features extracted from layer \(l\) of a pre-trained network, \(H_l\) and \(W_l\) are the height and width of the feature map at layer \(l\), \(\mathbf{w}_l\) is a learned weight vector specific to layer \(l\), and \(\odot\) denotes element-wise multiplication.

In our study, we utilize \ac{LPIPS} with AlexNet \cite{10.5555/2999134.2999257} as the feature extraction network to quantify the perceptual differences between the original images and their corresponding occluded versions. Specifically, we compute \ac{LPIPS} scores for image patches, comparing the original patches with their reconstructions generated through various occlusion strategies. These \ac{LPIPS} scores are then directly employed as an \ac{OoD} detection metric. This methodology is based on the premise that perceptual metrics like \ac{LPIPS} are particularly effective for inpainting tasks, as they are sensitive to subtle changes and anomalies, making them useful for identifying \ac{OoD} samples. The detailed results of our \ac{LPIPS} analysis are presented in Table \ref{table1}.

\begin{table}[h]
\caption{\ac{LPIPS} Scores for different occlusion strategies. Each $\text{Mask}_i$ corresponds to the inpainting of specific regions in the image, progressively targeting areas of decreasing importance as identified by the heatmap.}
\centering
\begin{tabular}{lcccc}
\hline
\toprule
\rowcolor{gray!20}
\textbf{Occlusion Strategy} & \textbf{$\text{Mask}_1$} & \textbf{$\text{Mask}_2$} & \textbf{$\text{Mask}_3$} & \textbf{$\text{Mask}_4$} \\
\bottomrule
\hline
Blackening & 0.1106 & 0.2280 & 0.3693 & 0.5567 \\
Histogram  & 0.0841 & 0.1844 & 0.3082 & 0.4621 \\
Mean  & 0.0796 & 0.1776 & 0.2997 & 0.4520 \\
\ac{NLI}  & 0.0781 & 0.1769 & 0.2999 & 0.4537 \\
Blurring  & 0.0701 & 0.1593 & 0.2670 & 0.3895 \\
\textbf{\ac{IBO}}  & \textbf{0.0381} & \textbf{0.0826} & \textbf{0.1407} & \textbf{0.2180} \\
\hline
\end{tabular}
    \label{table1}
\end{table}

\subsection{Proposed framework quantitative evaluation}

To evaluate our proposed occlusion approach, as shown in Table \ref{table2}, we directly calculated the \ac{IoU} between the ground-truth binary mask, indicating tumor regions in each sample, and the heatmap generated by the various \ac{XAI} techniques. This process allowed us to create a ground-truth ranking for all \ac{XAI} methods by averaging the \ac{IoU} values across a test set of samples. 
The Intersection over Union \ac{IoU} is defined as:

\begin{equation}
\text{\ac{IoU}} = \frac{R_{\text{ht}} \cap R_{\text{gt}}}{R_{\text{ht}} \cup R_{\text{gt}}}
\end{equation}

Where \(R_{\text{ht}}\) represents the region indicated by the heatmap, and \(R_{\text{gt}}\) denotes the ground truth region.

\begin{table}[h]
   \caption{\ac{IoU} between ground-truth and heatmaps generated by \ac{CAM}-based approaches. This ranking was used as ground-truth to evaluate occlusion strategies.}
    \centering
    \begin{tabular}{l c}
        \toprule
        \rowcolor{gray!20} 
        \textbf{Approach} & \textbf{\ac{IoU}} \\
        \midrule
        Full-Grad        & 0.6896 \\ 
        Grad-CAM         & 0.6482 \\ 
        Grad-CAM++       & 0.6467 \\ 
        XGrad-CAM        & 0.6464 \\ 
        Score-CAM        & 0.6452 \\ 
        Ablation-CAM     & 0.6439 \\ 
        Eigen-CAM        & 0.4257 \\ 
        \bottomrule
    \end{tabular}
    \label{table2}
\end{table}

To evaluate the effectiveness of heatmaps generated by \ac{CAM}-based interpretability methods, we use the \ac{AUC} to assess how well each heatmap correlates with the model’s tumor classification probability. Specifically, as we progressively occlude regions identified by the heatmaps as important, we track changes in the predicted probability of the tumor class. The \ac{AUC} of this relationship measures the heatmap's effectiveness.

In each occlusion step, we calculate the ratio of occluded regions to all identified regions, where a ratio of 0 means no regions are occluded and 1 indicates that all critical regions have been removed. A smaller \ac{AUC} suggests that the occluded regions were crucial for tumor classification, implying that the heatmap effectively highlighted significant tumor-related areas. By comparing \ac{AUC} values across different heatmap methods, we can identify which method most accurately captures critical regions relevant to the model’s decision-making.

For occlusion-based methods, including \ac{IBO}, the \ac{AUC} quantifies how the model's tumor classification probability changes as progressively larger regions, identified as important by the heatmaps, are occluded. The \ac{AUC} is defined as:

\begin{equation}
\text{\ac{AUC}} = \int_{0}^{1} f(p) \, dp
\end{equation}

where \( f(p) \) represents the tumor classification probability as a function of the percentage of occluded pixels, with \( p \) normalized to the range \([0, 1]\). Here, \( p \) ranges from 0 to 1, representing 0\% to 100\% occlusion, and \( f(p) \) denotes the classification probability for the tumor given the percentage \( p \) of occluded pixels.

A lower \ac{AUC} indicates that the occluded regions were critical to the model's decision, signifying accurate heatmap highlighting. We compare \ac{AUC} values from different heatmap methods to determine which technique best identifies key regions influencing the model’s tumor classification. Additionally, we compare the \ac{AUC} rankings with those obtained using the \ac{IoU} metric to find the most reliable occlusion method. The \ac{AUC} values for various methods, including our \ac{IBO} approach, are presented in Tables \ref{table3}--\ref{table8}.

\noindent
\begin{minipage}[t]{0.32\textwidth}
    \captionof{table}{\ac{IBO} rankings}
    \centering
\captionsetup{font=footnotesize}
    \begin{tabular}{l c}
        \toprule
        \rowcolor{gray!20} \textbf{Approach} & \textbf{\ac{AUC}
        } \\
        \midrule
        \textbf{Full-Grad} & 0.5335 \\
        \textbf{Grad-CAM} & 0.5991 \\
        \textbf{Grad-CAM++} & 0.6014 \\
        \textbf{XGrad-CAM} & 0.6058 \\
        Ablation-CAM & 0.6363 \\
        Score-CAM & 0.6707 \\
        \textbf{Eigen-CAM} & 0.8622 \\
        \bottomrule
    \end{tabular}
    \label{table3}
\end{minipage}%
\hfill
\begin{minipage}[t]{0.32\textwidth}
    \captionof{table}{Blurring rankings}
    \centering
    \captionsetup{font=footnotesize}
    \begin{tabular}{l c}
        \toprule
        \rowcolor{gray!20} \textbf{Approach} & \textbf{\ac{AUC}} \\
        \midrule
        \textbf{Full-Grad} & 0.4769 \\
        \textbf{Grad-CAM} & 0.5360 \\
        XGrad-CAM & 0.5360 \\
        Ablation-CAM & 0.5651 \\
        Grad-CAM++ & 0.5898 \\
        Score-CAM & 0.6094 \\
        \textbf{Eigen-CAM} & 0.8511 \\
        \bottomrule
    \end{tabular}
    \label{table4}
\end{minipage}%
\hfill
\begin{minipage}[t]{0.32\textwidth}
    \captionof{table}{\ac{NLI} rankings}
    \centering
    \captionsetup{font=footnotesize}
    \begin{tabular}{l c}
        \toprule
        \rowcolor{gray!20} \textbf{Approach} & \textbf{\ac{AUC}} \\
        \midrule
        \textbf{Full-Grad} & 0.3788 \\
        \textbf{Grad-CAM} & 0.4265 \\
        XGrad-CAM & 0.4278 \\
        Ablation-CAM & 0.4612 \\
        Grad-CAM++ & 0.4850 \\
        Score-CAM & 0.5154 \\
        \textbf{Eigen-CAM} & 0.7916 \\
        \bottomrule
    \end{tabular}
    \label{table5}
\end{minipage}
\vspace{10pt} 

\noindent
\begin{minipage}[t]{0.32\textwidth}
    \captionof{table}{Histogram rankings}
    \centering
    \captionsetup{font=footnotesize}
    \begin{tabular}{l c}
        \toprule
        \rowcolor{gray!20} \textbf{Approach} & \textbf{\ac{AUC}} \\
        \midrule
        \textbf{Full-Grad} & 0.3823 \\
        \textbf{Grad-CAM} & 0.4316 \\
        XGrad-CAM & 0.4335 \\
        Ablation-CAM & 0.4634 \\
        Grad-CAM++ & 0.4887 \\
        Score-CAM & 0.5214 \\
        \textbf{Eigen-CAM} & 0.7899 \\
        \bottomrule
    \end{tabular}
    \label{table6}
\end{minipage}%
\hfill
\begin{minipage}[t]
{0.32\textwidth}
    \captionof{table}{Mean rankings}
    \centering
\captionsetup{font=footnotesize}
    \begin{tabular}{l c}
        \toprule
        \rowcolor{gray!20} \textbf{Approach} & \textbf{\ac{AUC}} \\
        \midrule
        \textbf{Full-Grad} & 0.3718 \\
        XGrad-CAM & 0.4224 \\
        Grad-CAM & 0.4233 \\
        Ablation-CAM & 0.4537 \\
        Grad-CAM++ & 0.4790 \\
        Score-CAM & 0.5078 \\
        \textbf{Eigen-CAM} & 0.7871 \\
        \bottomrule
    \end{tabular}
    \label{table7}
\end{minipage}%
\hfill
\begin{minipage}[t]{0.32\textwidth}
    \captionof{table}{Blackening rankings}
    \centering
\captionsetup{font=footnotesize}
    \begin{tabular}{l c}
        \toprule
        \rowcolor{gray!20} \textbf{Approach} & \textbf{\ac{AUC}} \\
        \midrule
        \textbf{Full-Grad} & 0.4857 \\
        Grad-CAM++ & 0.5594 \\
        Grad-CAM  & 0.5624 \\
        Score-CAM & 0.5897 \\
        Ablation-CAM & 0.6061 \\
        XGrad-CAM & 0.6290 \\
        \textbf{Eigen-CAM} & 0.8630 \\
        \bottomrule
    \end{tabular}
    \label{table8}
\end{minipage}

Our approach demonstrates greater consistency in evaluating \ac{XAI} methods compared to other occlusion-based strategies. Notably, five out of the seven \ac{XAI} methods are ranked in the same order as in the \ac{IoU} table, indicating that our method is superior for removing features depicted in heatmaps. This results in a 71\% accuracy in predicting their performance outcomes, compared to 42\% with the best-performing alternative, making our approach a more reliable tool for evaluating \ac{XAI} methods. In addition, we evaluate the ranking performance of different methods by computing the \ac{MARD}. \ac{MARD} is a metric that quantifies the average deviation between the ranks assigned by a specific occlusion strategy and the ground-truth ranks. Mathematically, \ac{MARD} is defined as follows:

\begin{equation}
\text{MARD} = \frac{1}{N} \sum_{i=1}^{N} \left| \text{Rank}_{\text{GT}}(i) - \text{Rank}_{\text{oc}}(i) \right|
\end{equation}

where \(N\) is the total number of \ac{CAM}-based approaches, \(\text{Rank}_{\text{GT}}(i)\) denotes the ground-truth rank of \ac{CAM}-based approach \(i\), and \(\text{Rank}_{\text{oc}}(i)\) represents the rank assigned by a particular occlusion strategy. To apply this metric, we first established a ground-truth ranking of methods based on predefined criteria. Subsequently, we compared this ground-truth ranking to the rankings generated by various occlusion strategies. The goal was to measure how closely each strategy's ranking aligns with the ground-truth ranking. The results of this analysis are summarized in Table \ref{tab:MARD}. The table presents the \ac{MARD} values for each method, ranked from the highest to the lowest. A lower \ac{MARD} value indicates a closer alignment with the ground-truth ranking, suggesting a better performance.

\begin{table}[h]
\caption{\ac{MARD} values for occlusion strategies.}
\centering
\begin{tabular}{ll}
\toprule
\rowcolor{gray!20}
\textbf{Table} & \textbf{\ac{MARD} Value} \\
\midrule
Mean & 1.1428 \\
Blackening & 0.8571 \\
Histogram & 0.8571 \\
\ac{NLI} & 0.8571 \\
Blurring & 0.8571 \\
\textbf{\ac{IBO}} & \textbf{0.2857} \\
\bottomrule
\end{tabular}
\label{tab:MARD}
\end{table}

\section{Limitations and future work}
\label{future}

While our proposed \ac{IBO} strategy offers significant improvements in the evaluation of \ac{XAI} methods by reducing the generation of \ac{OoD} samples, it is not without limitations. One limitation of our \ac{IBO} strategy is that, despite utilizing a \ac{DDPM}, the inpainting process might not always maintain the full semantic integrity of the inpainted regions. The model, while effective in generating realistic non-cancerous tissue, may occasionally fail to capture the intricate details and variability of actual cancerous regions, which could affect the accuracy of the evaluation. Future work should focus on enhancing the \ac{DDPM} or integrating advanced generative models that are better suited for preserving the complex features of histopathological images, thereby improving the fidelity of the inpainting process.

Another limitation lies in the computational intensity of the \ac{IBO} strategy. The process of inpainting, particularly when applied iteratively across multiple masks, can be computationally expensive and time-consuming, which may not be practical for large-scale datasets or real-time applications. To address this, future research should explore optimizing the inpainting process for efficiency, potentially by developing more streamlined algorithms or leveraging hardware acceleration techniques such as GPU optimization, to make the approach more feasible for broader use.

Finally, our method primarily focuses on evaluating the impact of inpainting on tumor classification without fully exploring its effects on other potential applications within histopathological analysis, such as segmentation or multi-class classification. This narrow focus might limit the applicability of the method to broader tasks within medical imaging. Future work should aim to extend the \ac{IBO} strategy to encompass a wider range of applications, investigating how inpainting can influence other key aspects of histopathological image analysis, and thus provide a more comprehensive evaluation framework for \ac{XAI} methods.
\section{Conclusion}
\label{Conclusion}

In this paper, we proposed a novel \ac{IBO} strategy aimed at improving the evaluation of \ac{XAI} methods in histopathological image analysis. Our approach utilizes a \ac{DDPM} to inpaint occluded regions with realistic, non-cancerous tissue, effectively reducing the generation of \ac{OoD} samples that commonly arise in traditional occlusion strategies. This enhancement ensures that the evaluation of \ac{XAI} methods remains accurate and reliable by maintaining the integrity of the original data distribution.

Our experiments on the CAMELYON16 dataset employed a two-phase evaluation approach. In the first phase, we used the \ac{LPIPS} metric to measure the perceptual similarity between original and inpainted images, with lower \ac{LPIPS} scores indicating lower \ac{OoD} artifacts. Our \ac{IBO} method achieved significantly lower \ac{LPIPS} scores compared to other occlusion strategies, suggesting it better preserves the original tissue characteristics. In the second phase, we established a ground truth ranking of \ac{XAI} methods using \ac{IoU} scores and then evaluated these methods with the \ac{IBO} strategy and other approaches. By calculating the \ac{MARD}, we found that the \ac{IBO} method provided rankings much closer to the ground truth, demonstrating its superior accuracy in comparing \ac{XAI} methods.

In conclusion, the \ac{IBO} strategy not only enhances the interpretability of deep learning models in histopathology but also sets a new standard for the evaluation of \ac{XAI} methods by improving the accuracy of effectiveness rankings. The ability of \ac{IBO} to generate more realistic inpainted regions and provide more reliable \ac{XAI} evaluations underscores its potential for broader application in medical imaging.
\section*{List of Acronyms}
\begin{acronym}[VBLAST]
    \acro{XAI}{Explainable Artificial Intelligence}
    \acro{AI}{Artificial Intelligence}
    \acro{NLI}{Noisy Linear Imputation}
    \acro{CNN}{Convolutional Neural Network}
    \acro{OoD}{Out-of-Distribution}
    \acro{WSI}{whole slide image}
    \acro{AUC}{Area Under the Curve}
    \acro{CAM}{Class Activation Mapping}
    \acro{DDPM}{Denoising Diffusion Probabilistic Model}
    \acro{IoU}{Intersection over Union}
    \acro{IBO}{Inpainting-Based Occlusion}
    \acro{LPIPS}{Learned Perceptual Image Patch Similarity}
    \acro{MARD}{Mean Absolute Rank Difference}
    \acro{XML}{Extensible Markup Language}
    \acro{WCSS}{Within-Cluster Sum of Squares}
\end{acronym}
\bibliographystyle{unsrtnat}
\bibliography{references}

\begin{thebibliography}{48}
\providecommand{\natexlab}[1]{#1}
\providecommand{\url}[1]{\texttt{#1}}
\expandafter\ifx\csname urlstyle\endcsname\relax
  \providecommand{\doi}[1]{doi: #1}\else
  \providecommand{\doi}{doi: \begingroup \urlstyle{rm}\Url}\fi

\bibitem[Bayramoglu et~al.(2016)Bayramoglu, Kannala, and Heikkilä]{7900002}
Neslihan Bayramoglu, Juho Kannala, and Janne Heikkilä.
\newblock Deep learning for magnification independent breast cancer histopathology image classification.
\newblock In \emph{2016 23rd International Conference on Pattern Recognition (ICPR)}, pages 2440--2445, 2016.
\newblock \doi{10.1109/ICPR.2016.7900002}.

\bibitem[Dimitriou et~al.(2019)Dimitriou, Arandjelovi{\'c}, and Caie]{dimitriou2019deep}
Neofytos Dimitriou, Ognjen Arandjelovi{\'c}, and Peter~D Caie.
\newblock Deep learning for whole slide image analysis: an overview.
\newblock \emph{Frontiers in medicine}, 6:\penalty0 264, 2019.

\bibitem[Beyer et~al.(2023)Beyer, Rabiee, Ostadabbas, and Abiri]{beyer2023human}
Robert Beyer, Ali Rabiee, Sarah Ostadabbas, and Reza Abiri.
\newblock Human-inspired vision-based reaching and grasping for assistive robotic arms with reinforcement learning.
\newblock In \emph{society for neuroscience}, 2023.

\bibitem[Das et~al.(2020)Das, Conjeti, Chatterjee, and Sheet]{9269335}
Kausik Das, Sailesh Conjeti, Jyotirmoy Chatterjee, and Debdoot Sheet.
\newblock Detection of breast cancer from whole slide histopathological images using deep multiple instance cnn.
\newblock \emph{IEEE Access}, 8:\penalty0 213502--213511, 2020.
\newblock \doi{10.1109/ACCESS.2020.3040106}.

\bibitem[Lucieri et~al.(2020)Lucieri, Bajwa, Dengel, and Ahmed]{lucieri2020achievements}
Adriano Lucieri, Muhammad~Naseer Bajwa, Andreas Dengel, and Sheraz Ahmed.
\newblock Achievements and challenges in explaining deep learning based computer-aided diagnosis systems.
\newblock \emph{arXiv preprint arXiv:2011.13169}, 2020.

\bibitem[Abdelsamea et~al.(2022)Abdelsamea, Zidan, Senousy, Gaber, Rakha, and Ilyas]{https://doi.org/10.1002/widm.1474}
Mohammed~M. Abdelsamea, Usama Zidan, Zakaria Senousy, Mohamed~Medhat Gaber, Emad Rakha, and Mohammad Ilyas.
\newblock A survey on artificial intelligence in histopathology image analysis.
\newblock \emph{WIREs Data Mining and Knowledge Discovery}, 12\penalty0 (6):\penalty0 e1474, 2022.
\newblock \doi{https://doi.org/10.1002/widm.1474}.
\newblock URL \url{https://wires.onlinelibrary.wiley.com/doi/abs/10.1002/widm.1474}.

\bibitem[Khani et~al.(2024)Khani, Moeinaddini, Abnavi, and Shahraki]{khani2024explainable}
Pouya Khani, Elham Moeinaddini, Narges~Dehghan Abnavi, and Amin Shahraki.
\newblock Explainable artificial intelligence for feature selection in network traffic classification: A comparative study.
\newblock \emph{Transactions on Emerging Telecommunications Technologies}, 35\penalty0 (4):\penalty0 e4970, 2024.

\bibitem[Aminimehr et~al.(2024)Aminimehr, Khani, Molaei, Kazemeini, and Cambria]{aminimehr2024tbexplain}
Amirhossein Aminimehr, Pouya Khani, Amirali Molaei, Amirmohammad Kazemeini, and Erik Cambria.
\newblock Tbexplain: A text-based explanation method for scene classification models with the statistical prediction correction.
\newblock In \emph{Proceedings of the Conference on Governance, Understanding and Integration of Data for Effective and Responsible AI}, pages 54--60, 2024.

\bibitem[Rabiee et~al.(2024)Rabiee, Ghafoori, Cetera, and Abiri]{rabiee2024wavelet}
Ali Rabiee, Sima Ghafoori, Anna Cetera, and Reza Abiri.
\newblock Wavelet analysis of noninvasive eeg signals discriminates complex and natural grasp types.
\newblock \emph{arXiv preprint arXiv:2402.09447}, 2024.

\bibitem[Tosun et~al.(2020)Tosun, Pullara, Becich, Taylor, Chennubhotla, and Fine]{Tosun2020}
Akif~Burak Tosun, Filippo Pullara, Michael~J. Becich, D.~Lansing Taylor, S.~Chakra Chennubhotla, and Jeffrey~L. Fine.
\newblock \emph{HistoMapr{\texttrademark}: An Explainable AI (xAI) Platform for Computational Pathology Solutions}, pages 204--227.
\newblock Springer International Publishing, Cham, 2020.
\newblock ISBN 978-3-030-50402-1.
\newblock \doi{10.1007/978-3-030-50402-1_13}.
\newblock URL \url{https://doi.org/10.1007/978-3-030-50402-1_13}.

\bibitem[{van der Velden} et~al.(2022){van der Velden}, Kuijf, Gilhuijs, and Viergever]{VANDERVELDEN2022102470}
Bas~H.M. {van der Velden}, Hugo~J. Kuijf, Kenneth~G.A. Gilhuijs, and Max~A. Viergever.
\newblock Explainable artificial intelligence (xai) in deep learning-based medical image analysis.
\newblock \emph{Medical Image Analysis}, 79:\penalty0 102470, 2022.
\newblock ISSN 1361-8415.
\newblock \doi{https://doi.org/10.1016/j.media.2022.102470}.
\newblock URL \url{https://www.sciencedirect.com/science/article/pii/S1361841522001177}.

\bibitem[Lin et~al.(2021)Lin, Lee, and Celik]{lin2021you}
Yi-Shan Lin, Wen-Chuan Lee, and Z~Berkay Celik.
\newblock What do you see? evaluation of explainable artificial intelligence (xai) interpretability through neural backdoors.
\newblock In \emph{Proceedings of the 27th ACM SIGKDD conference on knowledge discovery \& data mining}, pages 1027--1035, 2021.

\bibitem[Buçinca et~al.(2020)Buçinca, Lin, Gajos, and Glassman]{Bu_inca_2020}
Zana Buçinca, Phoebe Lin, Krzysztof~Z. Gajos, and Elena~L. Glassman.
\newblock Proxy tasks and subjective measures can be misleading in evaluating explainable ai systems.
\newblock In \emph{Proceedings of the 25th International Conference on Intelligent User Interfaces}, IUI ’20. ACM, March 2020.
\newblock \doi{10.1145/3377325.3377498}.
\newblock URL \url{http://dx.doi.org/10.1145/3377325.3377498}.

\bibitem[Tomsett et~al.(2020)Tomsett, Harborne, Chakraborty, Gurram, and Preece]{tomsett2020sanity}
Richard Tomsett, Dan Harborne, Supriyo Chakraborty, Prudhvi Gurram, and Alun Preece.
\newblock Sanity checks for saliency metrics.
\newblock In \emph{Proceedings of the AAAI conference on artificial intelligence}, volume~34, pages 6021--6029, 2020.

\bibitem[Yeh et~al.(2019)Yeh, Hsieh, Suggala, Inouye, and Ravikumar]{yeh2019infidelitysensitivityexplanations}
Chih-Kuan Yeh, Cheng-Yu Hsieh, Arun~Sai Suggala, David~I. Inouye, and Pradeep Ravikumar.
\newblock On the (in)fidelity and sensitivity for explanations, 2019.
\newblock URL \url{https://arxiv.org/abs/1901.09392}.

\bibitem[Zeiler and Fergus(2014)]{zeiler2014visualizing}
Matthew~D Zeiler and Rob Fergus.
\newblock Visualizing and understanding convolutional networks.
\newblock \emph{arXiv preprint arXiv:1311.2901}, 2014.

\bibitem[Ivanovs et~al.(2021)Ivanovs, Kadikis, and Ozols]{IVANOVS2021228}
Maksims Ivanovs, Roberts Kadikis, and Kaspars Ozols.
\newblock Perturbation-based methods for explaining deep neural networks: A survey.
\newblock \emph{Pattern Recognition Letters}, 150:\penalty0 228--234, 2021.
\newblock ISSN 0167-8655.
\newblock \doi{https://doi.org/10.1016/j.patrec.2021.06.030}.
\newblock URL \url{https://www.sciencedirect.com/science/article/pii/S0167865521002440}.

\bibitem[Samek et~al.(2016)Samek, Binder, Montavon, Lapuschkin, and M{\"u}ller]{samek2016evaluating}
Wojciech Samek, Alexander Binder, Gr{\'e}goire Montavon, Sebastian Lapuschkin, and Klaus-Robert M{\"u}ller.
\newblock Evaluating the visualization of what a deep neural network has learned.
\newblock \emph{IEEE transactions on neural networks and learning systems}, 28\penalty0 (11):\penalty0 2660--2673, 2016.

\bibitem[Miró-Nicolau et~al.(2024)Miró-Nicolau, i~Capó, and Moyà-Alcover]{mirónicolau2024comprehensivestudyfidelitymetrics}
Miquel Miró-Nicolau, Antoni~Jaume i~Capó, and Gabriel Moyà-Alcover.
\newblock A comprehensive study on fidelity metrics for xai, 2024.
\newblock URL \url{https://arxiv.org/abs/2401.10640}.

\bibitem[Gomez et~al.(2022)Gomez, Fréour, and Mouchère]{gomez2022metricssaliencymapevaluation}
Tristan Gomez, Thomas Fréour, and Harold Mouchère.
\newblock Metrics for saliency map evaluation of deep learning explanation methods, 2022.
\newblock URL \url{https://arxiv.org/abs/2201.13291}.

\bibitem[Kindermans et~al.(2019)]{kindermans2019unreliability}
Pieter-Jan Kindermans et~al.
\newblock The (un)reliability of saliency methods.
\newblock In \emph{Explainable AI: Interpreting, Explaining and Visualizing Deep Learning}, pages 267--280. Springer, 2019.

\bibitem[Dasgupta et~al.(2022)Dasgupta, Frost, and Moshkovitz]{dasgupta2022framework}
Sanjoy Dasgupta, Nathan Frost, and Martin Moshkovitz.
\newblock Framework for evaluating faithfulness of local explanations.
\newblock In \emph{Proceedings of the 39th International Conference on Machine Learning}, pages 4794--4815. PMLR, 2022.

\bibitem[Nauta et~al.(2023)]{nauta2023systematic}
Meike Nauta et~al.
\newblock From anecdotal evidence to quantitative evaluation methods: a systematic review on evaluating explainable ai.
\newblock \emph{ACM Computing Surveys (CSUR)}, 55\penalty0 (13s):\penalty0 1--42, 2023.

\bibitem[Lamprou et~al.(2024)Lamprou, Kallipolitis, and Maglogiannis]{lamprou2024evaluation}
Vangelis Lamprou, Athanasios Kallipolitis, and Ilias Maglogiannis.
\newblock On the evaluation of deep learning interpretability methods for medical images under the scope of faithfulness.
\newblock \emph{Computer Methods and Programs in Biomedicine}, page 108238, 2024.

\bibitem[Rajaraman et~al.(2019)Rajaraman, Candemir, Thoma, and Antani]{rajaraman2019visualizing}
S~Rajaraman, S~Candemir, G~Thoma, and S~Antani.
\newblock Visualizing and explaining deep learning predictions for pneumonia detection in pediatric chest radiographs.
\newblock In \emph{Medical Imaging: Computer-Aided Diagnosis}, volume 10950. SPIE, 2019.

\bibitem[Sayres et~al.(2019)]{sayres2019using}
Rory Sayres et~al.
\newblock Using a deep learning algorithm and integrated gradients explanation to assist grading for diabetic retinopathy.
\newblock \emph{Ophthalmology}, 126\penalty0 (4):\penalty0 552--564, 2019.

\bibitem[Sundararajan et~al.(2017)Sundararajan, Taly, and Yan]{sundararajan2017axiomaticattributiondeepnetworks}
Mukund Sundararajan, Ankur Taly, and Qiqi Yan.
\newblock Axiomatic attribution for deep networks, 2017.
\newblock URL \url{https://arxiv.org/abs/1703.01365}.

\bibitem[Graziani et~al.(2021)Graziani, Lompech, M{\"u}ller, and Andrearczyk]{graziani2021evaluation}
Mara Graziani, Thomas Lompech, Henning M{\"u}ller, and Vincent Andrearczyk.
\newblock Evaluation and comparison of cnn visual explanations for histopathology.
\newblock In \emph{Proceedings of the AAAI Conference on Artificial Intelligence Workshops (XAI-AAAI-21), Virtual Event}, pages 8--9, 2021.

\bibitem[Tomsett et~al.(2019)Tomsett, Harborne, Chakraborty, Gurram, and Preece]{tomsett2019sanitycheckssaliencymetrics}
Richard Tomsett, Dan Harborne, Supriyo Chakraborty, Prudhvi Gurram, and Alun Preece.
\newblock Sanity checks for saliency metrics, 2019.
\newblock URL \url{https://arxiv.org/abs/1912.01451}.

\bibitem[Fong and Vedaldi(2017)]{Fong_2017}
Ruth~C. Fong and Andrea Vedaldi.
\newblock Interpretable explanations of black boxes by meaningful perturbation.
\newblock In \emph{2017 IEEE International Conference on Computer Vision (ICCV)}. IEEE, October 2017.
\newblock \doi{10.1109/iccv.2017.371}.
\newblock URL \url{http://dx.doi.org/10.1109/ICCV.2017.371}.

\bibitem[Wei et~al.(2018)Wei, Chang, Ying, Lim, and Lyu]{8546302}
Yi~Wei, Ming-Ching Chang, Yiming Ying, Ser~Nam Lim, and Siwei Lyu.
\newblock Explain black-box image classifications using superpixel-based interpretation.
\newblock In \emph{2018 24th International Conference on Pattern Recognition (ICPR)}, pages 1640--1645, 2018.
\newblock \doi{10.1109/ICPR.2018.8546302}.

\bibitem[Rong et~al.(2022)Rong, Leemann, Borisov, Kasneci, and Kasneci]{rong2022consistent}
Yao Rong, Tobias Leemann, Vadim Borisov, Gjergji Kasneci, and Enkelejda Kasneci.
\newblock A consistent and efficient evaluation strategy for attribution methods.
\newblock \emph{arXiv preprint arXiv:2202.00449}, 2022.

\bibitem[Ho et~al.(2020)Ho, Jain, and Abbeel]{ho2020denoisingdiffusionprobabilisticmodels}
Jonathan Ho, Ajay Jain, and Pieter Abbeel.
\newblock Denoising diffusion probabilistic models, 2020.
\newblock URL \url{https://arxiv.org/abs/2006.11239}.

\bibitem[Lugmayr et~al.(2022)Lugmayr, Danelljan, Romero, Yu, Timofte, and Gool]{lugmayr2022repaintinpaintingusingdenoising}
Andreas Lugmayr, Martin Danelljan, Andres Romero, Fisher Yu, Radu Timofte, and Luc~Van Gool.
\newblock Repaint: Inpainting using denoising diffusion probabilistic models, 2022.
\newblock URL \url{https://arxiv.org/abs/2201.09865}.

\bibitem[Oquab et~al.(2015)Oquab, Bottou, Laptev, and Sivic]{Oquab_2015_CVPR}
Maxime Oquab, Leon Bottou, Ivan Laptev, and Josef Sivic.
\newblock Is object localization for free? - weakly-supervised learning with convolutional neural networks.
\newblock In \emph{Proceedings of the IEEE Conference on Computer Vision and Pattern Recognition (CVPR)}, June 2015.

\bibitem[Hartigan and Wong(1979)]{Hartigan1979}
J.~A. Hartigan and M.~A. Wong.
\newblock A k-means clustering algorithm.
\newblock \emph{JSTOR: Applied Statistics}, 28\penalty0 (1):\penalty0 100--108, 1979.

\bibitem[Eht(2017)]{Ehteshami2017Bejnordi}
{Diagnostic Assessment of Deep Learning Algorithms for Detection of Lymph Node Metastases in Women With Breast Cancer}.
\newblock \emph{JAMA}, 318\penalty0 (22):\penalty0 2199--2210, 2017.

\bibitem[Roy et~al.(2021)Roy, Panda, and Jangid]{9616117}
Santanu Roy, Shubhajit Panda, and Mahesh Jangid.
\newblock Modified reinhard algorithm for color normalization of colorectal cancer histopathology images.
\newblock In \emph{2021 29th European Signal Processing Conference (EUSIPCO)}, pages 1231--1235, 2021.
\newblock \doi{10.23919/EUSIPCO54536.2021.9616117}.

\bibitem[He et~al.(2015)He, Zhang, Ren, and Sun]{he2015deepresiduallearningimage}
Kaiming He, Xiangyu Zhang, Shaoqing Ren, and Jian Sun.
\newblock Deep residual learning for image recognition, 2015.
\newblock URL \url{https://arxiv.org/abs/1512.03385}.

\bibitem[Selvaraju et~al.(2019)Selvaraju, Cogswell, Das, Vedantam, Parikh, and Batra]{Selvaraju_2019}
Ramprasaath~R. Selvaraju, Michael Cogswell, Abhishek Das, Ramakrishna Vedantam, Devi Parikh, and Dhruv Batra.
\newblock Grad-cam: Visual explanations from deep networks via gradient-based localization.
\newblock \emph{International Journal of Computer Vision}, 128\penalty0 (2):\penalty0 336–359, October 2019.
\newblock ISSN 1573-1405.
\newblock \doi{10.1007/s11263-019-01228-7}.
\newblock URL \url{http://dx.doi.org/10.1007/s11263-019-01228-7}.

\bibitem[Chattopadhay et~al.(2018)Chattopadhay, Sarkar, Howlader, and Balasubramanian]{8354201}
Aditya Chattopadhay, Anirban Sarkar, Prantik Howlader, and Vineeth~N Balasubramanian.
\newblock Grad-cam++: Generalized gradient-based visual explanations for deep convolutional networks.
\newblock In \emph{2018 IEEE Winter Conference on Applications of Computer Vision (WACV)}, pages 839--847, 2018.
\newblock \doi{10.1109/WACV.2018.00097}.

\bibitem[Fu et~al.(2020)Fu, Hu, Dong, Guo, Gao, and Li]{fu2020axiombasedgradcamaccuratevisualization}
Ruigang Fu, Qingyong Hu, Xiaohu Dong, Yulan Guo, Yinghui Gao, and Biao Li.
\newblock Axiom-based grad-cam: Towards accurate visualization and explanation of cnns, 2020.
\newblock URL \url{https://arxiv.org/abs/2008.02312}.

\bibitem[Desai and Ramaswamy(2020)]{9093360}
Saurabh Desai and Harish~G. Ramaswamy.
\newblock Ablation-cam: Visual explanations for deep convolutional network via gradient-free localization.
\newblock In \emph{2020 IEEE Winter Conference on Applications of Computer Vision (WACV)}, pages 972--980, 2020.
\newblock \doi{10.1109/WACV45572.2020.9093360}.

\bibitem[Muhammad and Yeasin(2020)]{Muhammad_2020}
Mohammed~Bany Muhammad and Mohammed Yeasin.
\newblock Eigen-cam: Class activation map using principal components.
\newblock In \emph{2020 International Joint Conference on Neural Networks (IJCNN)}. IEEE, July 2020.
\newblock \doi{10.1109/ijcnn48605.2020.9206626}.
\newblock URL \url{http://dx.doi.org/10.1109/IJCNN48605.2020.9206626}.

\bibitem[Wang et~al.(2020)Wang, Wang, Du, Yang, Zhang, Ding, Mardziel, and Hu]{wang2020scorecamscoreweightedvisualexplanations}
Haofan Wang, Zifan Wang, Mengnan Du, Fan Yang, Zijian Zhang, Sirui Ding, Piotr Mardziel, and Xia Hu.
\newblock Score-cam: Score-weighted visual explanations for convolutional neural networks, 2020.
\newblock URL \url{https://arxiv.org/abs/1910.01279}.

\bibitem[Srinivas and Fleuret(2019)]{srinivas2019fullgradientrepresentationneuralnetwork}
Suraj Srinivas and Francois Fleuret.
\newblock Full-gradient representation for neural network visualization, 2019.
\newblock URL \url{https://arxiv.org/abs/1905.00780}.

\bibitem[Zhang et~al.(2018)Zhang, Isola, Efros, Shechtman, and Wang]{zhang2018unreasonableeffectivenessdeepfeatures}
Richard Zhang, Phillip Isola, Alexei~A. Efros, Eli Shechtman, and Oliver Wang.
\newblock The unreasonable effectiveness of deep features as a perceptual metric, 2018.
\newblock URL \url{https://arxiv.org/abs/1801.03924}.

\bibitem[Krizhevsky et~al.(2012)Krizhevsky, Sutskever, and Hinton]{10.5555/2999134.2999257}
Alex Krizhevsky, Ilya Sutskever, and Geoffrey~E. Hinton.
\newblock Imagenet classification with deep convolutional neural networks.
\newblock In \emph{Proceedings of the 25th International Conference on Neural Information Processing Systems - Volume 1}, NIPS'12, page 1097–1105, Red Hook, NY, USA, 2012. Curran Associates Inc.

\end{thebibliography}

\end{document}